\documentclass[letterpaper, 10 pt, conference]{ieeeconf}  

\IEEEoverridecommandlockouts                              

\usepackage{graphicx} 
\usepackage{amsmath} 
\usepackage{amssymb}  
\usepackage{algorithm}
\usepackage{algorithmic} 
\usepackage[T1]{fontenc}
\usepackage{multicol}
\usepackage{multirow}
\usepackage{subfigure}
\usepackage{color}
\usepackage{cite}
\usepackage{placeins}
\usepackage{booktabs}
\usepackage{balance}
\usepackage{caption}

\usepackage{epstopdf}
\AppendGraphicsExtensions{.tif}

\usepackage{hyperref}
\usepackage{cleveref}
\hypersetup{hidelinks}

\title{\LARGE \bf
DexGraspNet: A Large-Scale Robotic Dexterous\\ Grasp Dataset for General Objects Based on Simulation}

\author{Ruicheng Wang$^{1*}$, Jialiang Zhang$^{1*}$, Jiayi Chen$^{1,2}$, Yinzhen Xu$^{1,2}$, Puhao Li$^{2,3}$, Tengyu Liu$^{2}$, He Wang$^{1\dagger}$ 
\thanks{$^{1}$Peking University}
\thanks{$^{2}$Beijing Institute for General Artificial Intelligence}
\thanks{$^{3}$Tsinghua University}
\thanks{$^{*}$Equal contribution}
\thanks{$^{\dagger}$Corresponding author: \tt\small hewang@pku.edu.cn}
}

\begin{document}

\maketitle
\thispagestyle{empty}
\pagestyle{empty}

\begin{abstract}
Robotic dexterous grasping is the first step to 
enable human-like dexterous object manipulation and thus a crucial robotic technology.
However, dexterous grasping is much more under-explored than object grasping with parallel grippers, partially due to the lack of a large-scale dataset.
In this work, we present a large-scale robotic dexterous grasp dataset, DexGraspNet, 
generated by our proposed highly efficient synthesis method that can be generally applied to any dexterous hand.
Our method leverages a deeply accelerated differentiable force closure estimator and thus can efficiently and robustly synthesize stable and diverse grasps on a large scale.
We choose ShadowHand and generate 1.32 million grasps for 5355 objects, covering more than 133 object categories and containing more than 200 diverse grasps for each object instance, with all grasps having been validated by the Isaac Gym simulator. 
Compared to the previous dataset from Liu et al. generated by \textit{GraspIt!}, our dataset has not only more objects and grasps, but also higher diversity and quality.
Via performing cross-dataset experiments, we show that training several algorithms of dexterous grasp synthesis on our dataset significantly outperforms training on the previous one.
To access our data and code, including code for human and Allegro grasp synthesis, please visit our project page: \href{https://pku-epic.github.io/DexGraspNet/}{https://pku-epic.github.io/DexGraspNet/}. 

\end{abstract}


\section{INTRODUCTION}

Robotic object grasping is an important technology for many robotic systems. Recent years have witnessed great success in developing vision-based grasping methods~\cite{breyer2021volumetric, sundermeyer2021contact, fang2020graspnet,gou2021RGB,Wang_2021_ICCV,fang2022transcg} and large-scale datasets for parallel-jaw grippers, \textit{e.g.}, synthetic object-centric dataset, ACRONYM~\cite{acronym2020}, and real-world dataset of grasping in clutter, GraspNet~\cite{fang2020graspnet}. 

\begin{figure}
    \centering
    \includegraphics[width=0.9\columnwidth]{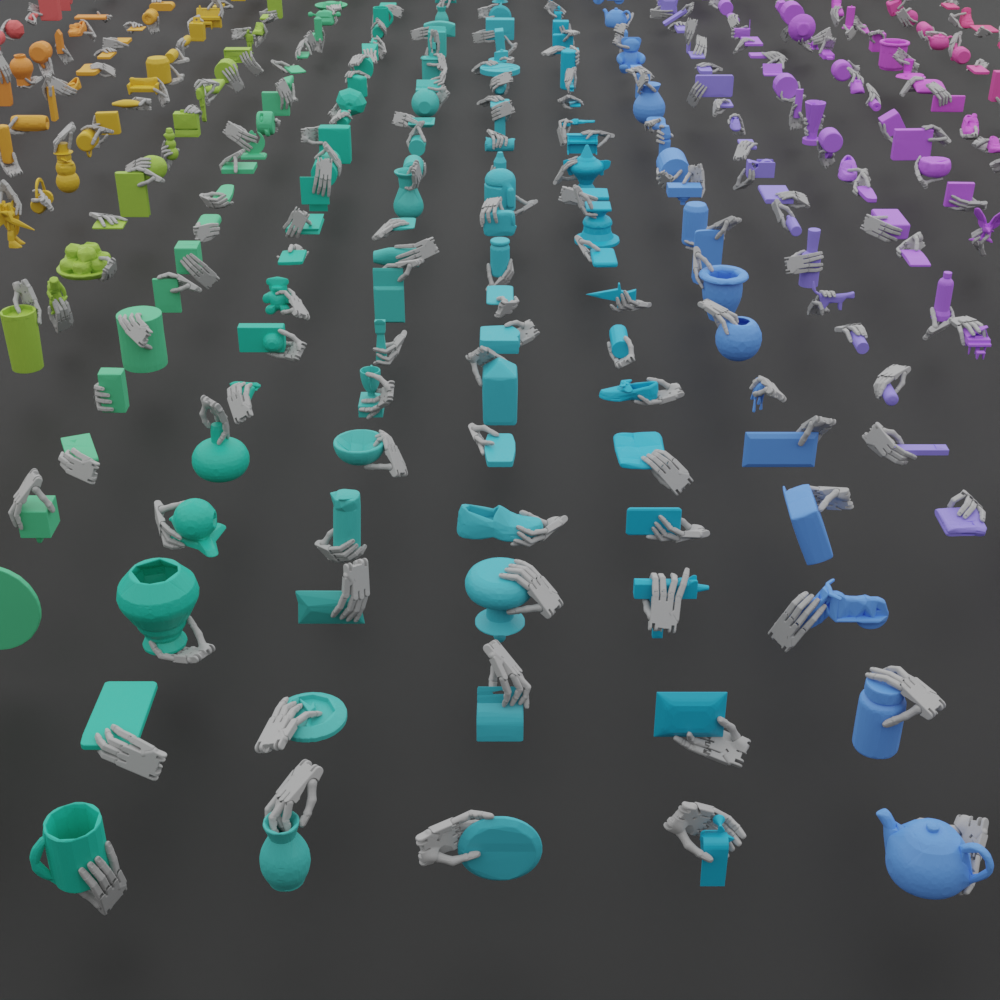}
    \caption{A visualization of DexGraspNet. DexGraspNet contains 1.32M grasps of ShadowHand\cite{shadowhand} on 5355 objects, which is two orders of magnitudes larger than the previous dataset from DDG~\cite{liu2020deep}. It features diverse types of grasping that cannot be achieved using \textit{GraspIt!}~\cite{graspit}.}
    \vspace{-12pt}
\end{figure}

Although simple and effective for pick-and-place, parallel-jaw grippers show certain limitations in dexterous object manipulation, \textit{e.g.}, using scissors, due to their low DoFs. 
On the contrary, multi-fingered robotic hands, \textit{e.g.}, ShadowHand~\cite{shadowhand}, are human-like, designed with very high DoFs (26 for ShadowHand), and can attain more diverse grasp types.
Those dexterous hands can support many complex and diverse manipulations, \textit{e.g.}, solving Rubik's cube~\cite{openaiRubikCube}, and can be used in task-specific grasping~\cite{kokic2017affordance}.

Arguably, dexterous grasping is the first step to dexterous manipulation. However, dexterous grasping is highly under-explored, compared to parallel grasping.
One major obstacle is the lack of large-scale robotic dexterous  grasping datasets required by learning-based methods.
Up to now, the only dataset is provided by Liu et al.~\cite{liu2020deep} (Deep Differentiable Grasp, referred to as DDG), which contains only 6.9K grasps and 565 objects and is much smaller than the grasp datasets for parallel grippers, \textit{e.g.}, GraspNet~\cite{fang2020graspnet}, ACRONYM~\cite{acronym2020}. 
Considering the high-DoF nature of the dexterous hand, dexterous grasping datasets need to be significantly larger and more diverse for the sake of generalization. 

In this work, we propose DexGraspNet, a large-scale simulated dataset for robotic dexterous grasping. This dataset contains 1.32 million dexterous grasps for ShadowHand on 5355 objects, with more than 200 diverse grasps for each object instance. 
The objects are from more than 133 hand-scale object categories and collected from various synthetic and scanned object datasets.
In addition to the scale, our dataset also features high diversity and high physical stability. All grasps have been examined by force closure and validated by Isaac Gym~\cite{isaacGym} physics simulator, enabling further tasks in both real-world and simulation environments.

\begin{table*}[tb]
		\setlength{\abovecaptionskip}{0cm}
		\setlength{\belowcaptionskip}{0.2cm}
	\centering
	\caption{\label{tab:l1} Dexterous Grasp Dataset Comparison} 
    \begin{tabular}{c|c|c|c|c|c|c|c}
        Dataset & Hand & Observations & Sim./Real & Grasps & Obj.(Cat.) & Grasps per Obj. & Method \\
        \hline\hline
        ObMan~\cite{obman} & MANO & - & Sim. & 27k & 2772(8) & 10 & \textit{GraspIt!}  \\
        HO3D~\cite{ho3d} & MANO & RGBD & Real & 77k & 10 & >7k & Estimation \\
        DexYCB~\cite{DexYCB} & MANO & RGBD & Real & 582K & 20 & \textbf{>29k} & Human annotation \\
        ContactDB~\cite{contactdb} & MANO & RGBD+thermal & Real & 3750 & 50 & 75 & Capture \\
        ContactPose~\cite{contactpose} & MANO & RGBD & Real & 2306 & 25 & ~92 & Capture\\
        DDGdata~\cite{liu2020deep} & ShadowHand & - & Sim. & 6.9k & 565 & $>$100 & \textit{GraspIt!} \\
        DexGraspNet (Ours) & ShadowHand & - & Sim. & \textbf{1.32M} & \textbf{5355(133)} & $>$200 & Optimization \\
    \end{tabular}
    \vspace{-12pt}
\end{table*}

Note that synthesizing diverse high-quality dexterous grasps at scale is known to be very challenging. 
For dexterous grasping data synthesis, previous works, \textit{e.g.}, DDG, mainly use \textit{GraspIt!}~\cite{graspit}, which lacks diversity in grasping poses due to its naive search strategy. 
A recent work \cite{liu2021synthesizing} proposes a novel method to address this diversity issue. This work devises a differentiable energy term to approximate force closure and then uses it to synthesize diverse and stable grasps via optimization. However, \cite{liu2021synthesizing} suffers from low yield, slow convergence, and strict constraints on object meshes, making it infeasible for us to use for synthesizing a large-scale dataset. 

To achieve our desired diversity, quality, and scale, we propose several critical improvements to~\cite{liu2021synthesizing}, making it much more efficient and robust. 
First, we design a better hand pose initialization strategy and carefully select contact candidates to boost yield. For synthesizing 10000 valid grasps, we speed up from 400 GPU hours to 7 GPU hours. 
Second, we propose an alternative way to compute penetration energy and signed distances, which enables us to handle object meshes of much lower quality, and also highly simplifies their preprocessing procedures. 
Third, we introduce energy terms that punish self-penetration and out-of-limit joint angles to further improve grasp quality. 
Additionally, with simple modifications, the entire pipeline can be applied to other dexterous hands, such as MANO \cite{MANO} and Allegro.

To verify the advantage of our dataset over the one from DDG, we train two dexterous grasping algorithms on our dataset and DDG. The cross-dataset experiments confirm that training on our dataset yields better grasping quality and higher diversity. Also, the great diversity of the hand grasps from our dataset leaves huge improvement space for future dexterous grasping algorithms.

\section{RELATED WORK}

Researches in grasping can be broadly categorized by the types of end effectors involved. The most thoroughly studied ones are the suction cup and parallel jaw grippers, whose grasp pose can be defined by a 7D vector at most, including 3D for translation, 3D for rotation, and 1D for the width between the two fingers. Dexterous robotic hands with three or more fingers such as ShadowHand~\cite{shadowhand} and humanoid hands such as MANO~\cite{MANO} require more complex descriptors, sometimes up to 24DoF as in ShadowHand~\cite{shadowhand}. In this paper, we are dedicated to researches on the latter type. To bridge the gap between humanoid hands and robotic hands, numerous researches have shown the efficacy of retargeting humanoid hand poses to dexterous robotic hands~\cite{contactgrasp,mandikal2021learning,qin2022one,ye2022learning}. 

\subsection{Analytical Grasping}
Early researches in dexterous grasping focus on optimizing grasping poses to form force closure that can resist external forces and torques~\cite{rodriguez2012caging,prattichizzo2012manipulability,rosales2012synthesis,murray2017mathematical}. 

Due to the complexity of computing hand kinematics and testing force closure, many works were devoted to simplifying the search space~\cite{ponce1993characterizing,ponce1997computing,li2003computing}. As a result, these methods were applicable to restricted settings and can only produce limited types of grasping poses. Another stream of work~\cite{zheng2009distance,dai2018synthesis,liu1999qualitative} looks for simplifying the optimization process with an auxiliary function. \cite{liu2021synthesizing} proposed to use a differentiable estimator of the force closure metric to synthesize diverse grasping poses for arbitrary hands.

\subsection{Data-Driven Grasping}
Recent works shift their focus to data-driven methods. Given an object, the most straightforward approach is to directly generate the pose vectors of the grasping hand~\cite{grasptta,corona2020ganhand,multifingan,ddgc,yang2021cpf}. A refinement step is usually implemented in these methods to remove inconsistencies such as penetration. 

Other methods take an indirect approach that involves generating an intermediate representation first. Existing methods use contact points~\cite{unigrasp,wu2022learning,li2022efficientgrasp}, contact maps~\cite{varley2015generating,contactgrasp,turpin2022graspd,mandikal2021learning,zhu2021toward}, and occupancy fields~\cite{karunratanakul2020grasping} as the intermediate representations. The methods then obtain the grasping poses via optimization~\cite{unigrasp,wu2022learning,turpin2022graspd,karunratanakul2020grasping}, planning~\cite{varley2015generating}, RL policies~\cite{li2022efficientgrasp,mandikal2021learning}, or another generative model~\cite{zhu2021toward}. 

Compared to most analytical methods, data-driven methods show improved inference speed and diversity of generated grasping poses. However, the diversity is still limited by the training data.

\subsection{Dexterous Grasp Datasets}
Dexterous grasping is impossibly difficult to annotate for its overwhelming degrees of freedom. Most existing works are trained on programmatically synthesized grasping poses~\cite{goldfeder2009columbia,obman,ddgc,liu2020deep} using the \textit{GraspIt!}~\cite{graspit} planner. The planner first searches the eigengrasp space for pregrasp poses that cross a threshold. Then, the planner squeezes all fingers in the selected pregrasp poses to construct a firm grasp. Since the initial search is performed in the low dimensional eigengrasp space, the resulting data follows a narrow distribution and cannot cover the full dexterity of multi-finger hands.

More recent works leverage the increasing capacity of computer vision to collect human hand poses when interacting with the object. HO3D~\cite{ho3d,hampali2022keypointtransformer} computes the ground truth 3D hand pose for images from 2D hand keypoint annotations. The method resolves ambiguities by considering physics constraints in hand-object interactions and hand-hand interactions. DexYCB~\cite{DexYCB} and ContactPose~\cite{contactpose} solve the 3D hand shape from multi-view RGBD camera recordings. Latest datasets~\cite{GRAB,GOAL,ARCTIC} use optical motion capture systems to track hand and object shapes during interactions. While these methods produce natural and smooth interaction demonstrations, the data is restricted to humanoid hand structures and daily hand poses. 

In addition, ContactDB~\cite{contactdb} and ContactPose~\cite{contactpose} leverage IR cameras to collect contact maps on object surfaces. 

\section{DATASET GENERATION METHOD}

\subsection{Object Preperation}
We collect object models from various datasets. ShapeNetCore and ShapeNetSem~\cite{shapenet2015, shapenetsem} contain Computer-Aided-Design (CAD) models with category labels, from which we select 3980 objects in 133 categories. YCB~\cite{ycb}, BigBIRD~\cite{singh2014bigbird}, Grasp~\cite{graspDataset}, KIT~\cite{kasper2012kit}, and Google's scanned Object Dataset~\cite{downs2022google} are scanned model repositories without category labels, from which we select 1375 objects. They are not labeled with categories in our dataset.

Since not all objects from these datasets are aligned to sizes in the real world, we choose to normalize all models into a unit sphere and then augment each object by uniformly scaling them with 5 fixed sizes. Then we remesh them into manifolds~\cite{huang2020manifoldplus} to make closed figures. Finally, for simulation purposes, we create collision meshes for every object mesh through convex decomposition using CoACD~\cite{coacd}. 

\subsection{Grasp Generation}

\subsubsection{Assumptions and Notations} 
To parameterize dexterous grasps, we use tuples $g=(T, R, \theta)$, where $T\in \mathbb{R}^3$ and $R\in SO(3)$ form the global hand pose, and $\theta\in \mathbb{R}^{d}$ describes the $d$ joint angles ($d=22$ for ShadowHand). Given the URDF of the dexterous hand, we can use $g$ to compute the hand mesh $H$ via forward kinematics. 
In the generation process, we need to consider the contact points.
Following~\cite{liu2021synthesizing}'s idea that fewer contact candidates lead to faster convergence, we first manually select 140 contact candidates from the surface of the hand,  and $n$ contact points are randomly sampled among these candidates in each iteration of the optimization to calculate energy terms. We then augment our grasp tuple to $g'=(T,R,\theta,x)$, with $x$ representing the $n=4$ contact points. Given object mesh $O$, the contact normal vectors $c\in \mathbb{R}^{n\times 3}$ can be computed from $x$. Note that $x$ is an intermediate variable of the grasp generation algorithm and is discarded after the algorithm finishes.

\subsubsection{Review of Differentiable Force Closure}
Our dexterous grasp generation method is mainly built upon the original work~\cite{liu2021synthesizing}, in which they propose a novel differentiable force closure estimator as an energy term and use optimization to synthesize grasps. 
The proposed differentiable force closure term, which encourages a set of contact points to form force closure, can be expressed as
\begin{equation}
    E_{\rm fc}=\|Gc\|_2
\end{equation}
where
$$
G=\begin{bmatrix}
I_3&\cdots&I_3\\
[x_1]_\times&\cdots&[x_n]_\times
\end{bmatrix}
$$
$$
{[x_i]}_\times=\begin{bmatrix}0&-x_i^{(z)}&x_i^{(y)}\\x_i^{(z)}&0&-x_i^{(x)}\\-x_i^{(y)}&x_i^{(x)}&0\end{bmatrix}
$$
A pair of attraction and repulsion energy functions are introduced to ensure contact and prevent penetration:
\begin{equation}
    E_{\rm dis}=\sum _{i=1}^n d(x_n, O),  E_{\rm pen}=\sum_{v\in S(H)}[v\in O]d(v, O)   
\end{equation}
where $S(H)$ is the surface point cloud of the hand mesh $H$, $d(\cdot, \cdot)$ is the point-to-mesh distance and $[v\in O]=1$ if point $v$ is inside object mesh $O$. They also use an energy function $E_{\rm prior}$ to keep the hand in a natural state. The complete energy function is as follows:
\begin{equation}
    E=E_{\rm fc}+w_{\rm dis}E_{\rm dis}+w_{\rm pen}E_{\rm pen}+w_{\rm prior}E_{\rm prior}
\end{equation}
They design a modified MALA optimization algorithm to minimize the energy $E$ over the augmented grasp tuple $g'=(T,R,\theta,x)$. The algorithm takes an initial hand pose $g'_0$, which is randomly initialized. Then, in each iteration, $T,R,\theta$ are updated according to Langevin dynamics, and contact points $x$ are randomly re-sampled with a small probability. The update is accepted or rejected stochastically by the Metropolis-Hastings rule. The optimization ends after 10000 steps. For more details, please refer to~\cite{liu2021synthesizing}.

\subsubsection{Our Method}

Though~\cite{liu2021synthesizing} takes a great step forward, it is still quite hard to obtain a large-scale grasp dataset directly using their method due to three reasons: 1. the algorithm suffers from a low success rate and slow convergence; 2. most object meshes we use have no thickness due to the poor quality of the object dataset, making it impossible to compute penetration energy; 3. due to random initialization, some generated grasping poses may look twisted. To overcome these issues, we propose several ways to improve the efficiency, effectiveness, and robustness of the original algorithm:

First, we propose an initialization strategy that can greatly improve the success rate and speed up the convergence. We find that their optimization algorithm's success rate drops dramatically when the initial hand is closed. This motivates us to introduce two constraints to the initialization strategy: 1. the five fingers should be opened to form a space for grasping; 2. the palm should face the object. 

\begin{figure}
    \centering
        \begin{minipage}[b]{0.25\linewidth}
            \centerline{\includegraphics[width=\textwidth]{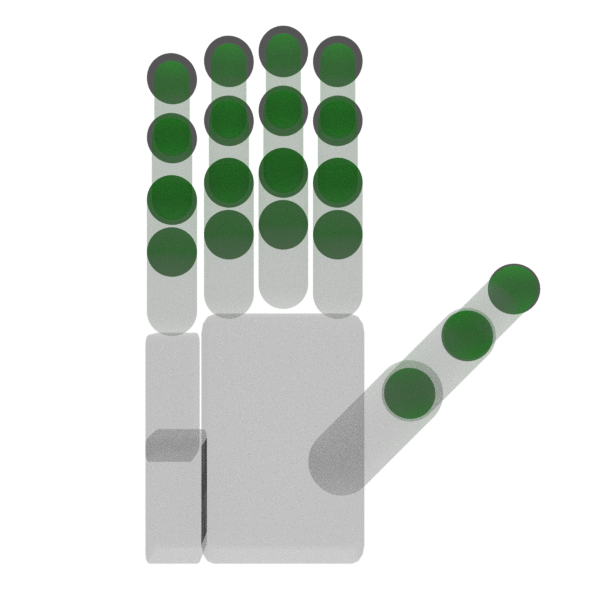}}
            \centerline{(a)}
        \end{minipage}
        \begin{minipage}[b]{0.25\linewidth}
            \centerline{\includegraphics[width=\textwidth]{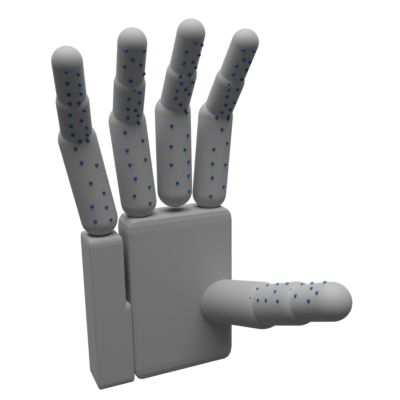}}
            \centerline{(b)}
        \end{minipage}
        \begin{minipage}[b]{0.35\linewidth}
            \centerline{\includegraphics[width=\textwidth]{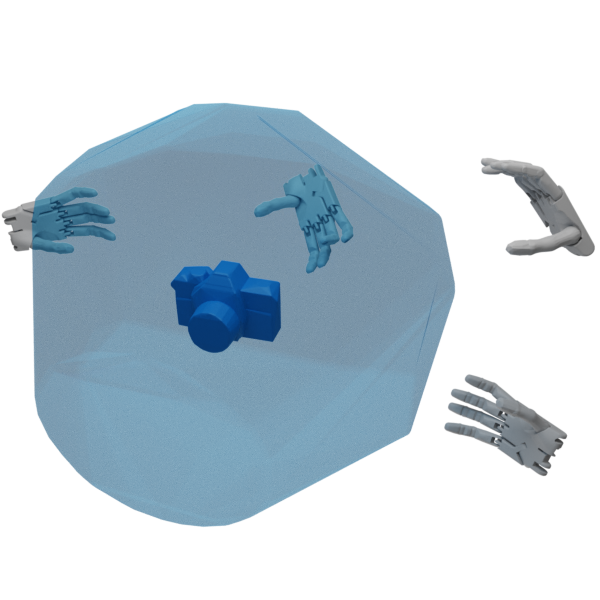}}
            \centerline{(c)}
        \end{minipage}
    \caption{(a) Green spheres with the radius of $\delta=1{\rm cm}$ are manually selected to compute $E_{\rm spen}$. (b) Contact candidates on the collision mesh in the canonical initial hand pose. 
    (c) Initialization: 1. sample points on the object's inflated convex hull (blue); 2. move the hands to the sampled points and jitter the translation, rotation, and joint angles.}
    \label{fig:initialization}
\end{figure}

More specifically, we manually choose a canonical hand pose $\theta_{\rm ref}$, as shown in \Cref{fig:initialization}, then jitter each joint angle within its limit using the truncated normal distribution. Then, on each object mesh, we first take its convex hull, then push every vertex of the hull away from the origin by $0.2m$ to obtain the inflated convex hull. Next, we sample a random point $p$ on the surface of the inflated convex hull, and compute the direction vector from $p$ to its nearest point on the original object mesh, then jitter this direction vector within a cone and get $\vec n$. Finally, the hand is moved to $p$, and rotated to face the same direction as $\vec n$, then push away from the object mesh along $\vec n$ by a random distance, and rotated around $\vec n$ randomly.

The resulting initial hand pose $(T_0,R_0,\theta_0)$ can be easily optimized to a grasp pose, thus raising the success rate. Moreover, for each object, if we sample enough initial hands, they can surround the object densely and evenly, so we can generate diverse data. Also, we found that using our strategy, the final grasp poses look more natural.

Second, we propose an alternative way to compute penetration energy that can make the algorithm more robust to thin object meshes of low quality. In practice, the original algorithm will fail completely when the object mesh has no thickness because the penetration energy will always be zero. Therefore, instead of taking the point cloud from the hand, we take it from the object and compute each point's distance to the hand mesh. We call this the reverse penetration energy. It does not require the object mesh to have any thickness at all, allowing us to process far more object CAD models. 

Third, we modify $E_{\rm prior}$, inspired by \cite{zhu2021toward} to penalize self penetration and out-of-limit joint angles:
\begin{equation}
    E_{\rm spen}=\sum_{p\in P(H)}\sum_{q\in P(H)}[p\ne q]\max(\delta-d(p,q), 0)
\end{equation}
\vspace{-5pt}
\begin{equation}
    E_{\rm joints}=\sum_{i=1}^d(\max(\theta_i-\theta_i^{\max},0)+\max(\theta_i^{\min}-\theta_i,0))
\end{equation}
These add to our final energy function:
\begin{equation}
    E_{\rm fc}+w_{\rm dis}E_{\rm dis}+w_{\rm pen}E_{\rm pen}+w_{\rm spen}E_{\rm spen}+w_{\rm joints}E_{\rm joints}
\end{equation}
where $w_{\rm dis}=100,w_{\rm pen}=100,w_{\rm spen}=10, w_{\rm joints}=1$.

Another minor difference between our implementation and~\cite{liu2021synthesizing} lies in the optimization algorithms. Because our initialization strategy has already raised the success rate of the algorithm to an acceptable level, we simplify MALA and use simple gradient descent to update $T,R,\theta$ during each optimization step. In our settings, the optimization process converges in less than 6000 iterations, which reduces almost half of the original iterations.

Finally, we use a modified version of Kaolin~\cite{jatavallabhula2019kaolin} instead of DeepSDF~\cite{park2019deepsdf} to compute point-to-mesh signed distances, which eliminates the need for pretraining category level DeepSDF networks, and significantly reduces the memory cost when optimizing grasps.

\begin{figure}
    \centering
    \includegraphics[width=1\columnwidth]{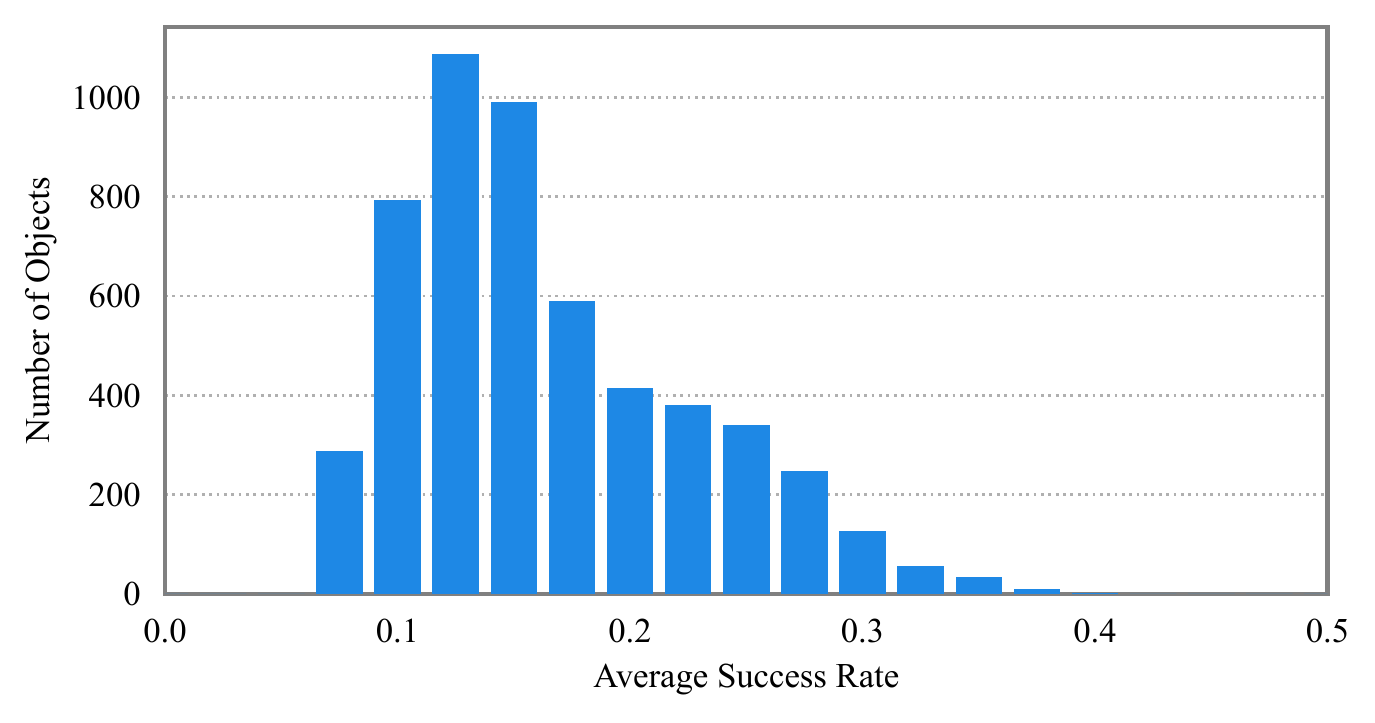}
    \caption{Distribution of object numbers with respect to the average success rate for each object after final validation. We only keep successful grasps in our dataset.}
    \label{fig:distribution_of_grasps_in_dataset}
\end{figure}

\begin{figure}
    \centering
        \begin{minipage}{0.8\linewidth}
            \centerline{
                \includegraphics[width=0.30\textwidth]{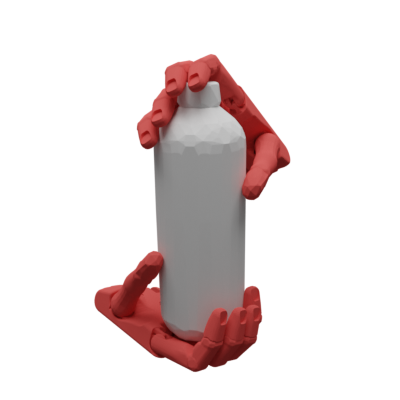}
                \includegraphics[width=0.30\textwidth]{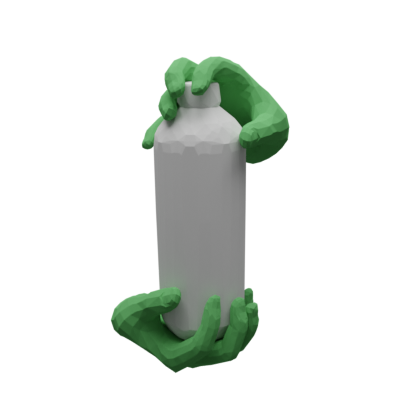}
                \includegraphics[width=0.30\textwidth]{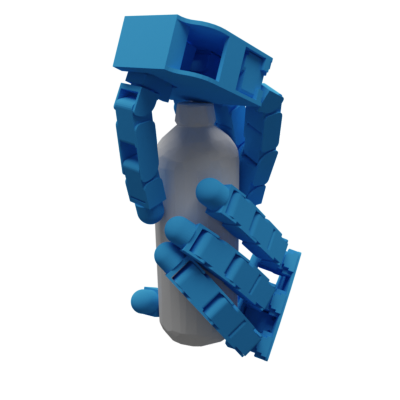}
            }
        \end{minipage}
        \begin{minipage}{0.8\linewidth}
            \centerline{
                \includegraphics[width=0.30\textwidth]{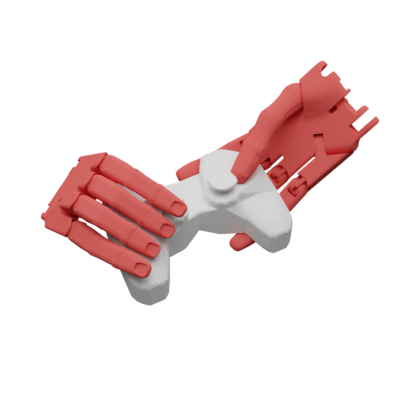}
                \includegraphics[width=0.30\textwidth]{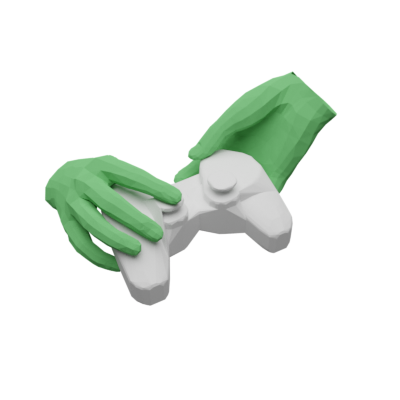}
                \includegraphics[width=0.30\textwidth]{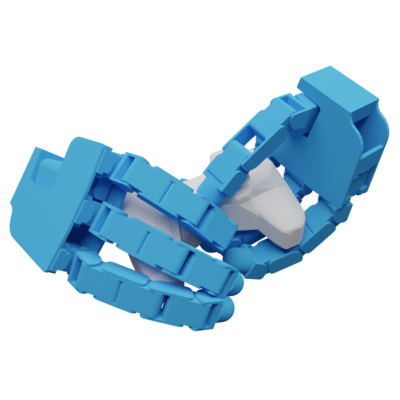}
            }
        \end{minipage}
    \caption{Visualization of grasps using different dexterous hands. From left to right: ShadowHand, MANO, Allegro.}
    \label{fig:three_hands}
\end{figure}

\subsection{Grasp Validation}
\label{sec: grasp validation}
To filter out those bad results after the optimization converges, we validate all of the grasps in a physical simulator Isaac Gym~\cite{isaacGym} with PhysX as the basic physics engine. 
We first initialize the gripper using the final grasp parameters. Then, in order to apply active forces on the object, we slightly move each contacting link of the gripper along the normal of its contact point, and set the moved pose as target positions for position control. Finally, gravity with a magnitude of $9.8 m/s^2$ is added to the scene. A grasp is considered successful if the gripper is still in contact with the object after 100 simulation steps under all 6 axis-aligned directions of gravity. The distribution of the object number with respect to the average success rate for each object is shown in Fig.~\ref{fig:distribution_of_grasps_in_dataset}.
Moreover, if the max penetration depth exceeds 0.1cm, we also consider the grasp as a failure. We only save those grasps who pass both the simulation validation and the penetration validation in our dataset.

\begin{figure*}[h!]
    \centering
    \begin{minipage}{0.95\linewidth}
            \centerline{
                \includegraphics[width=0.095\textwidth]{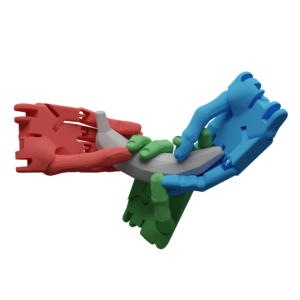}
                \includegraphics[width=0.095\textwidth]{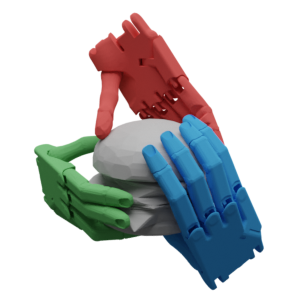}
                \includegraphics[width=0.095\textwidth]{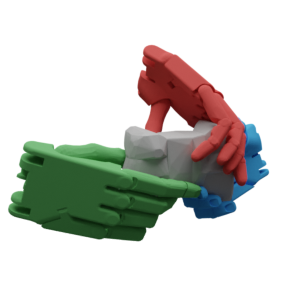}
                \includegraphics[width=0.095\textwidth]{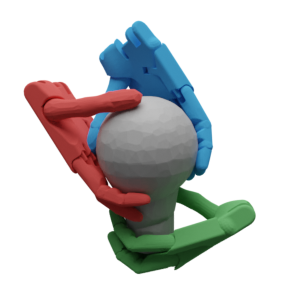}
                \includegraphics[width=0.095\textwidth]{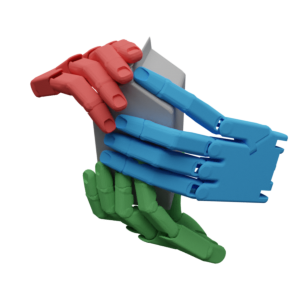}
                \includegraphics[width=0.095\textwidth]{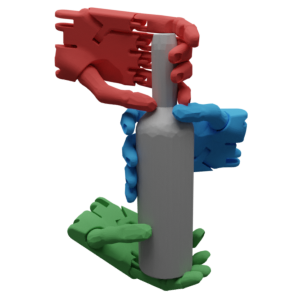}
                \includegraphics[width=0.095\textwidth]{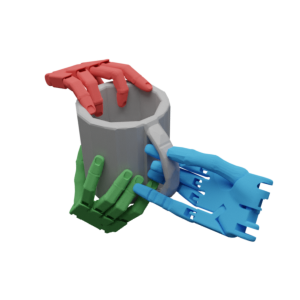}
                \includegraphics[width=0.095\textwidth]{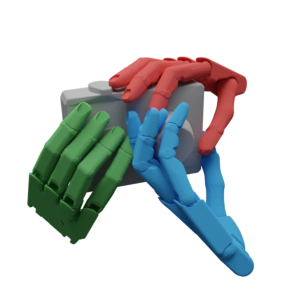}
                \includegraphics[width=0.095\textwidth]{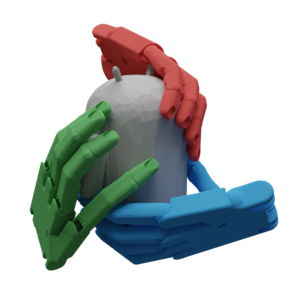}
                \includegraphics[width=0.095\textwidth]{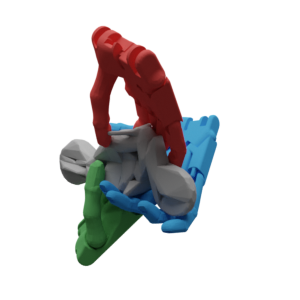}
            }
    \end{minipage}
\begin{minipage}{0.95\linewidth}
            \centerline{
                \includegraphics[width=0.095\textwidth]{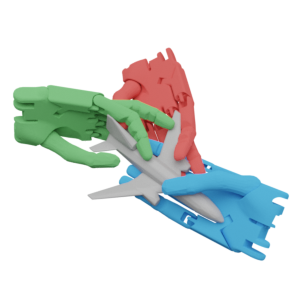}
                \includegraphics[width=0.095\textwidth]{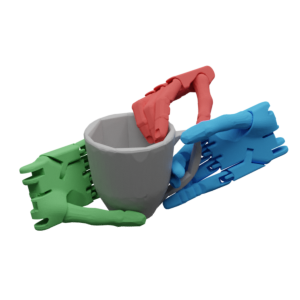}
                \includegraphics[width=0.095\textwidth]{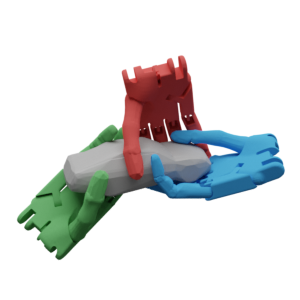}
                \includegraphics[width=0.095\textwidth]{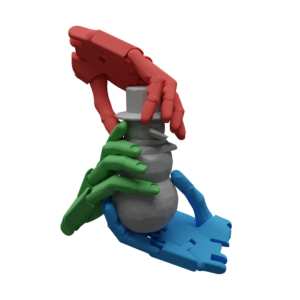}
                \includegraphics[width=0.095\textwidth]{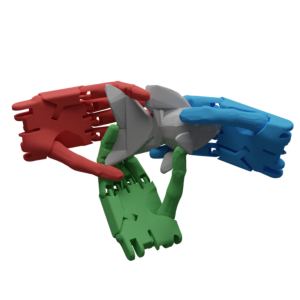}
                \includegraphics[width=0.095\textwidth]{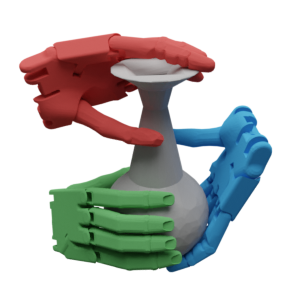}
                \includegraphics[width=0.095\textwidth]{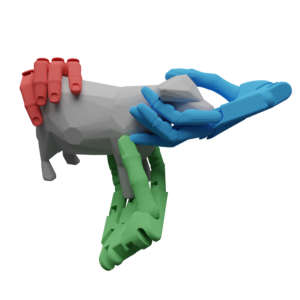}
                \includegraphics[width=0.095\textwidth]{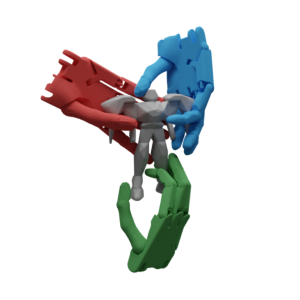}
                \includegraphics[width=0.095\textwidth]{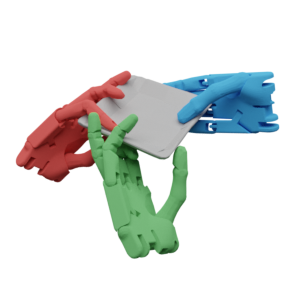}
                \includegraphics[width=0.095\textwidth]{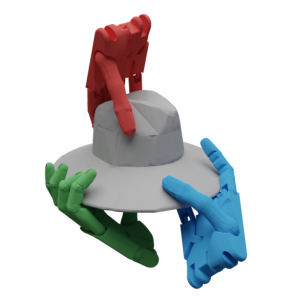}
        }
    \end{minipage}
    \caption{Visualization of the diverse grasps on the objects from DexGraspNet.}
    \label{fig:show_of_dgn}
\end{figure*}

\section{Dataset Analysis and Comparison}
\label{sec: dataset analysis}

With our improved pipeline, we generate more than 200 grasps per object, which sum up to 1.32 million grasps in total, forming the largest grasping dataset for ShadowHand. Some visual qualitative results are shown in Fig.~\ref{fig:show_of_dgn}. Additionally, this pipeline can be stably applied to other dexterous hands. Fig.~\ref{fig:three_hands} shows some synthesized results for human hand (MANO\cite{MANO}) and Allegro along with ShadowHand.

Compared to the original algorithm~\cite{liu2021synthesizing}, our improved pipeline achieves a significant speed-up. 
On NVIDIA A100 with 19.49 TFLOPS, our algorithm takes 74min to optimize 10000 grasps for 6000 steps, out of which about 18\% are considered valid under our settings. The original algorithm of \cite{liu2021synthesizing} takes 37min on NVIDIA 3090 with 35.58 TFLOPS to optimize 512 grasps for 10000 steps, out of which about 3\% are considered valid. 
It took us 950 GPU hours on A100 to generate 1.32 million valid grasps, which would have taken the original algorithm 50000 GPU hours. This speed-up is contributed by faster convergence, smaller memory (which leads to bigger batch size), and a higher success rate. 

We demonstrate the quality of our dataset by comparing DexGraspNet with the dataset proposed in DDG~\cite{liu2020deep} (in short DDGdata), a grasping dataset for ShadowHand generated by \textit{GraspIt!}~\cite{graspit}, on two aspects below. 

\begin{figure}[tb!]
    \centering
    \begin{minipage}{0.4\linewidth}
        \includegraphics[width=0.8\textwidth]{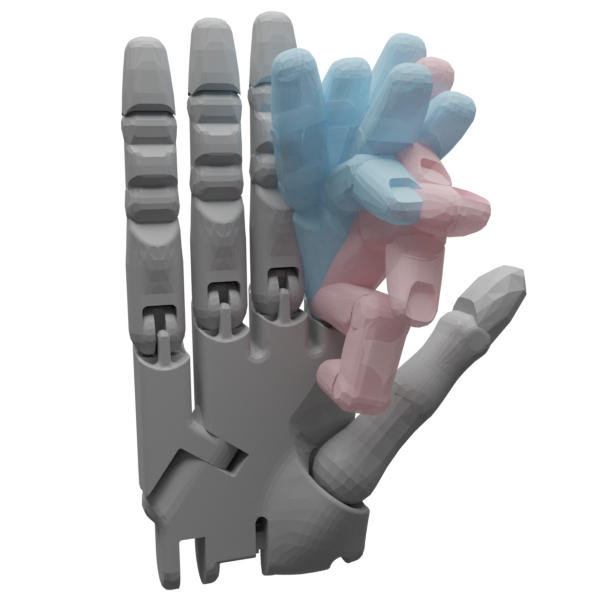}
    \end{minipage}
    \begin{minipage}{0.45\linewidth}
        \centerline{\includegraphics[width=\textwidth]{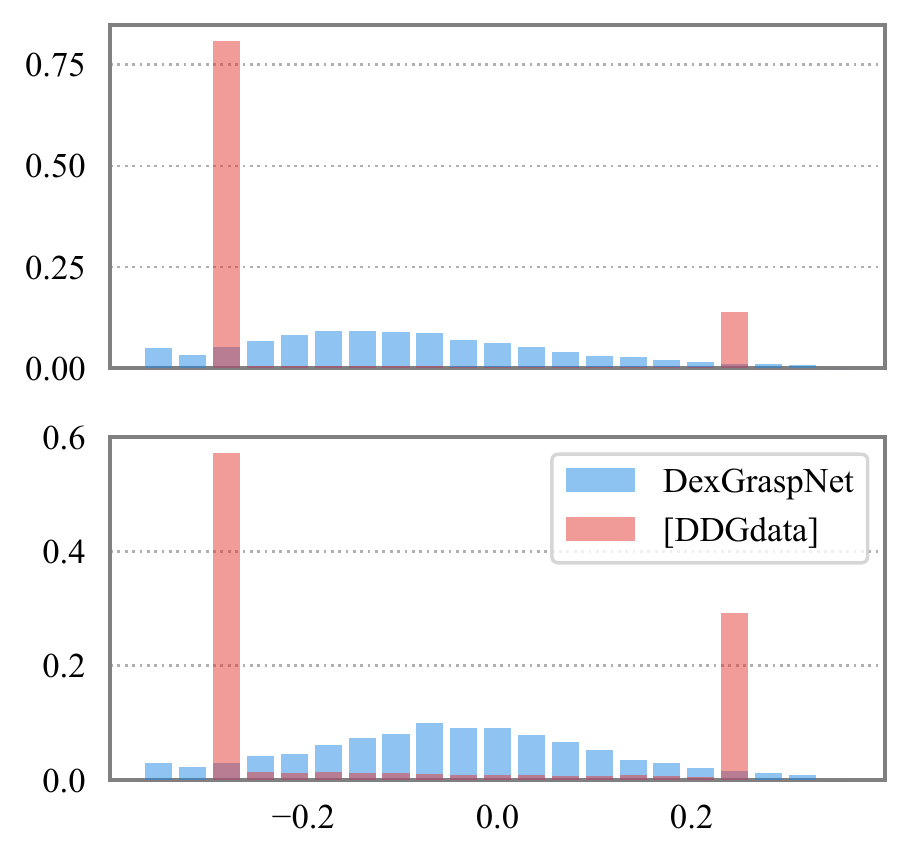}}
    \end{minipage}
    \caption{Diversity comparison between DexGraspNet and DDGdata. Left: The poses of each finger of DDGdata (red) all collapse into a fan-shaped space, while fingers from our data (blue) can reach out in all directions. Right: Probability distribution of two joint angles over all grasp poses. Joint angles from DDGdata typically hover around the two limits, contributing to the dataset's stiff and inflexible grasp mode. }
    \label{fig:mode_collapse}
\end{figure}

First, we conclude that DexGraspNet is more diverse. As shown in Fig.~\ref{fig:mode_collapse}, the planner in \textit{GraspIt!} can only clench each finger in a fixed direction, so the root joints of every finger lose one DoF each. Moreover, the angles of many joints in DDGdata often collapse to their upper or lower limits due to their simple generation strategy. These phenomena contribute to a serious loss of diversity. In contrast, our optimization method can generate diverse grasps with much higher dexterity, as shown in Fig.~\ref{fig:show_of_dgn}. We further use the mean entropy to model the diversity quantitatively.  
To evaluate this metric, we first discretize each joint's motion range into $100$ bins, then use samples from each dataset to estimate a probability distribution, calculate the entropy of these distributions, and take the mean over all joints.
Results are shown in Table~\ref{tab:diversity_analysis}. 

Second, we show that DexGraspNet is more stable by comparing the $Q_1$ metric~\cite{ferrari1992planning}, which is intuitively the norm of the smallest wrench that can destabilize the grasp:
\begin{equation}
    Q_1 = {\rm inscribed~sphere~radius~of~} {\rm ConvexHull}(\cup_i w_i)
\end{equation}
where $\{w_i\}$ are contact friction cone wrenches. We choose $1{\rm mm}$ as the contact threshold, and allow at most one contact point for each link to save computational time.  
The results are shown in Table~\ref{tab:diversity_analysis}. We can find that the average value of our dataset is significantly better than that of DDGdata. It is worth noting that our entire generation pipeline does not explicitly optimize these metrics.

\begin{table}[tb]
    \centering
    \caption{Statistics of $Q_1$ and Entropy}
    \begin{tabular}{c|ccc}
        Dataset & 100\% $Q_1$ mean& best 10\% $Q_1$ mean& $H$ mean\\
        \hline
        DDGdata & 0.0712  & 0.2277 & 4.246 \\
        DexGraspNet & \textbf{0.1145} & \textbf{0.2533} & \textbf{5.962}
    \end{tabular}
    \label{tab:diversity_analysis}
\end{table}

\section{Benchmarks}

\begin{table*}[t!]
    \centering
    \caption{Benchmarks of the Grasp Quality}
    \begin{tabular}{c|ccc|ccc}
        \multirow{2}{*}{Method (Training Dataset)} & \multicolumn{3}{c|}{Tested on DexGraspNet} & \multicolumn{3}{c}{Tested on DDGdata} \\
        \cline{2-7} 
        & success $\uparrow$ & $Q_1$ $\uparrow$ & pen $\downarrow$ & success $\uparrow$ & $Q_1$ $\uparrow$ & pen $\downarrow$ \\
        \hline\hline 
        DDG (DDGdata) & 57.4 & 0.0493 & 0.353 & 56.4 & 0.0461 & 0.333\\
        DDG (DexGraspNet) & \textbf{67.5} & \textbf{0.0582} & \textbf{0.173} & \textbf{75.9} & \textbf{0.0524} & \textbf{0.134}\\
        \hline
        GraspTTA (DDGdata) & 17.1 & 0.0126 & 0.720 & 23.7 & 0.0265 & 0.666\\
        GraspTTA (DexGraspNet) & \textbf{24.5} & \textbf{0.0271} & \textbf{0.678} & \textbf{39.3} & \textbf{0.0790} & \textbf{0.547}
    \end{tabular}
    \label{tab:benchmark}
\end{table*}

We benchmark two methods of 
dexterous grasp synthesis, DDG\cite{liu2020deep} and GraspTTA\cite{grasptta}, on our dataset, and compare them with the same methods trained on DDGdata\cite{liu2020deep}. 

\subsection{Benchmark Methods}

DDG \cite{liu2020deep} designs a differentiable $Q_1$ metric, which generalizes the standard $Q_1$ metric to the case when the gripper is not in contact with objects. With this generalized $Q_1$ metric, they are able to supervise the neural network to predict fine grasp end-to-end. Their network takes 5 depth images of the object as input and directly regresses 6D pose and joint angles of the ShadowHand. To ease learning, they divide the training process into two stages. In the first stage, they only use the loss of the grasp poses, and in the second stage, they fine-tune the network with differentiable $Q_1$ loss and other losses to avoid penetration and pull the hand closer to the object. We follow their data pre-process pipeline to generate BVH representations and depth images of our objects, and train the network with official settings on each dataset.

Another work GraspTTA (short for Test-Time Adaptation)~\cite{grasptta} proposes to synthesize high-quality grasps by ensuring the contact consistency between the hand and the object. They design two networks, one is a CVAE\cite{cvae} to synthesize grasps, and the other is a contact net to predict contact regions of the object. During training, those two networks are trained separately. During testing, a grasp is synthesized in a two-stage process. First, the CVAE takes the object point cloud as condition, samples a latent code, and then decodes the hand 6D global pose and joint angles, which can be further transformed into the hand point cloud through forward kinematics. Second, the contact net takes both the object and the hand point cloud to predict a target contact map, and optimizes the hand parameters to minimize the difference between the current contact map and the target contact map. We re-implement GraspTTA on ShadowHand, process the data from DexGraspNet and DDGdata in the same way as in~\cite{grasptta}, and train the networks on each dataset for the same number of iterations.

\subsection{Experiments and Results}

We report the following metrics for evaluation. 1)
\textbf{Simulation success rate(\%) in Isaac Gym}. We adopt an easier criterion (the criterion in Sec. \ref{sec: grasp validation} is too strict for the baselines): a grasp is considered valid if it can withstand at least one of the six gravity directions and has a maximal penetration less than $5{\rm mm}$. 2) \textbf{Mean} $Q_1$ \cite{ferrari1992planning}, which is introduced in Sec. \ref{sec: dataset analysis}. Since these methods cannot guarantee exact contact, we relax the contact threshold to $1{\rm cm}$. Particularly, if the penetration depth is greater than $5{\rm mm}$, the $Q_1$ metric is not well defined, so we manually set $Q_1$ of these results to 0. 3) \textbf{Maximal penetration depth(cm).} This is defined as the maximal penetration depth from the object point cloud to hand meshes.

The main results are presented in Table~\ref{tab:benchmark}.  Comparing models trained on DexGraspNet with models trained on DDGdata, we observe that no matter which baseline, test set, or metric we use, the former always scores higher. We thus conclude that learning-based grasping methods achieve higher performance when they are trained under our dataset. Table~\ref{tab:benchmark} also shows that the output of DDG has higher quality than GraspTTA most of the time. More specifically, GraspTTA suffers severely from penetration. 

Apart from grasp quality, we use joint angle entropy (the same as in Section IV) to evaluate the diversity of the grasps generated by the two methods. Table~\ref{tab:entropy_dataset} shows the joint angle entropy of models trained on DexGraspNet always have higher means and lower standard deviations than models trained on DDGdata, which means DexGraspNet improves the diversity of grasping methods. We also find that GraspTTA has higher diversity than DDG, partly due to the test time optimization used in GraspTTA that can generate many variations.
To compare the joint angle entropy of models trained on DexGraspNet and the original joint angle entropy of DexGraspNet, we further find that: 1) DDG's entropy is lower than DexGraspNet's entropy, meaning DDG cannot fully recover DexGraspNet's diversity; 2) although GraspTTA yields an entropy higher than DexGraspNet, this does not necessarily mean that GraspTTA learns diverse grasping, given its success rate is very low.
We interpret this as the trade-offs that DDG and GraspTTA individually make, given that stability and diversity are contradictory to some degree.
More importantly, this status quo shows that none of the existing grasping methods can fully learn the highly diverse grasp poses of DexGraspNet while keeping a reasonable success rate at the same time.

\begin{table}[t!]
    \centering
    \caption{Benchmarks of the Grasp Diversity}
    \begin{tabular}{c|cc|cc}
        \multirow{2}{*}{Method (Training Dataset)} & \multicolumn{2}{c|}{DexGraspNet} & \multicolumn{2}{c}{DDGdata}\\
        \cline{2-5}
        & $H$ mean&$H$ std & $H$ mean&$H$ std \\
        \hline\hline
        DDG (DDGdata)&4.958&2.653&3.709&1.942\\
        DDG (DexGraspNet)&\textbf{5.683}&\textbf{1.993}&\textbf{4.272}&\textbf{1.287}\\
        \hline
        GraspTTA (DDGdata)&5.952&0.934&5.837&1.047\\
        GraspTTA (DexGraspNet)&\textbf{6.111}&\textbf{0.569}&\textbf{5.947}&\textbf{0.528}
    \end{tabular}
    \label{tab:entropy_dataset}
\end{table}

\section{LIMITATIONS}

By comparing grasps in our dataset with the taxonomy from~\cite{feix2015grasp}, we notice that our dataset cannot cover every grasping type described. Since the optimization step tends to pull every candidate point closer to the object, the final grasps are always contact-rich, or power grasps. Therefore, precision grasps hardly appear, which represents the dexterity of multi-finger robotic hands.
Additionally, our method lacks semantic guidance, which makes it hard to generate functional grasps, \textit{e.g.} picking up the mug by its handle. Precision grasps and functional grasps remain important issues for us to explore.

\section{CONCLUSIONS}

In this paper, we present a large-scale synthetic dexterous grasping dataset, DexGraspNet, synthesized via our proposed deeply-accelerated optimization-based method.
This dataset has a much larger scale, better grasp quality, and higher diversity than previous datasets. Trained on DexGraspNet, previous grasp synthesis methods can achieve consistent improvements in both quality and diversity. However, none of the existing methods can perform well on both metrics.  
Compared to grasping using parallel grippers, we argue that dexterous grasping has a larger room for research. We release DexGraspNet and hope that its scale, quality, and diversity can help future methods tackle the task of dexterous grasping, and exploit more potential of dexterous grippers.





\section{Acknowledgements}

This work is supported in part by the National Key R\&D Program of China (2022ZD0114900) and the Beijing Municipal Science \& Technology Commission (Z221100003422004).





\bibliographystyle{IEEEtran}
\balance
\bibliography{reference}

\end{document}